\documentclass{amia}
\usepackage{lipsum} 
\usepackage{bm}
\usepackage{times}
\usepackage{helvet}
\usepackage{courier}
\usepackage{times}
\usepackage{latexsym}
\usepackage{amsmath}
\usepackage{algorithm}
\usepackage{algpseudocode}
\usepackage{url}
\usepackage{multirow}
\usepackage{enumitem}
\usepackage{array}
\usepackage[symbol]{footmisc}
\usepackage{threeparttable}
\usepackage{graphicx}
\usepackage{dblfloatfix} 
\usepackage{caption}
\usepackage[labelformat=simple]{subcaption}

\usepackage{adjustbox}
\usepackage{multirow}
\usepackage{hyperref}
\usepackage{amsmath}
\usepackage{amssymb}
\usepackage{subcaption}

\DeclareMathOperator*{\argmax}{argmax}
\newcolumntype{P}[1]{>{\centering\arraybackslash}p{#1}}

\begin{document}


\title{An Automatic SOAP Classification System Using Weakly Supervision And Transfer Learning}

\author{Sunjae Kwon, MSc$^1$, Zhichao Yang, MSc$^1$, Dr. Hong Yu, PhD$^{1, 2, 3, 4}$}

\institutes{
    $^1$ College of Information and Computer Science, University of Massachusetts Amherst, MA\\
    $^2$ Department of Medicine, University of Massachusetts Medical School, Worcester, MA\\
    $^3$ College of Information and Computer Science, University of Massachusetts Lowell, MA\\
    $^4$ Center for Healthcare Organization and Implementation Research, Bedford Veterans Affairs Medical Center, MA
}

\maketitle

\section*{Abstract}
\textbf{Objective:} In this paper, we introduce a comprehensive framework for developing a machine learning-based SOAP (Subjective, Objective, Assessment, and Plan) classification system without manually SOAP annotated training data or with less manually SOAP annotated training data.  

\textbf{Materials and Methods:} The system is composed of the following two parts: 1) Data construction, 2) A neural network-based SOAP classifier, and 3) Transfer learning framework. In data construction, since a manual construction of a large size training dataset is expensive, we propose a rule-based weak labeling method utilizing the structured information of an EHR note. Then, we present a SOAP classifier composed of a pre-trained language model and bi-directional long-short term memory with conditional random field (Bi-LSTM-CRF). Finally, we propose a transfer learning framework that re-uses the trained parameters of the SOAP classifier trained with the weakly labeled dataset for datasets collected from another hospital. 

\textbf{Results:} The proposed weakly label-based learning model successfully performed SOAP classification (89.99 F1-score) on the notes collected from the target hospital. Otherwise, in the notes collected from other hospitals and departments, the performance dramatically decreased. Meanwhile, we verified that the transfer learning framework is advantageous for inter-hospital adaptation of the model increasing the models' performance in every cases. In particular, the transfer learning approach was more efficient when the manually annotated data size was smaller. 

\textbf{Conclusion:} We showed that SOAP classification models trained with our weakly labeling algorithm can perform SOAP classification without manually annotated data on the EHR notes from the same hospital. The transfer learning framework helps SOAP classification model's inter-hospital migration with a minimal size of the manually annotated dataset.

\section{Introduction}
Electronic health record (EHR) notes are patient-centered clinical documentation that makes information securely accessible to authorized personnel. EHR notes are a fundamental resource for a broad range of clinical natural language processing (NLP) applications such as adverse drug detection\cite{jagannatha2019overview}, identification of associations between social determinants of health and opioid overdose\cite{mitra2021risk}, and predictions of diseases and outcomes\cite{liu2018deep, zhang2020combining}. 

EHR notes typically have a problem-oriented SOAP (Subjective, Objective, Assessment, and Plan) structure\cite{weed2014medical}. The $subjective$ section describes the patients’ current condition(s), which can be self-reported or the physician’s summary of previous status in the medical record relevant to the chief complaint. This section includes current symptoms, medical history, surgical history, family history, and social history along with current medications. The $objective$ section includes clinical conditions, laboratory, physical, and other examination findings from the current visit. The $assessment$ section typically contains medical diagnoses and summaries of the key elements that lead to the medical diagnoses. Following the diagnoses, physicians lay out the $plan$ for treatment or additional testing. 

Since each section of the SOAP structure describes the different clinical aspects of a patient, the SOAP structure information can play an important role in variety downstream biomedical natural language processing (BioNLP) systems that target to analyze EHR notes. Yang and Yu\cite{yang2020generating} suggest to utilize the SOAP structure of EHR notes for the automated clinical assessment generation (MCAG). Specifically, the information in the subjective and objective sections of the SOAP-structured EHR notes are used to train an automatic assessment generation model along with an additional medical knowledge graph. In addition, Gao et al.,\cite{gao2022hierarchical} introduces a new assessment and plan relation label task that identifies plan subsections relevant to the major diseases/problems from a given EHR note.

\begin{figure*}[t]
\centering
\begin{subfigure}[b]{.49\textwidth}
\includegraphics[width=\linewidth]{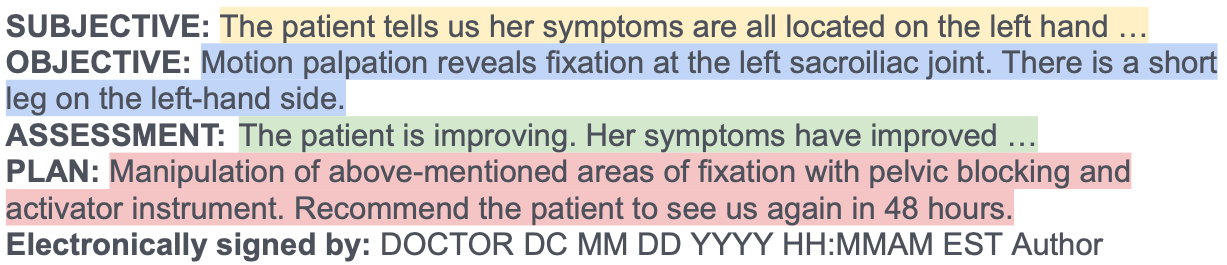}
\caption{An example with an explicitly structured SOAP note}
\label{fig:with_explicit}
\end{subfigure}
\hfill
\begin{subfigure}[b]{.49\textwidth}
\includegraphics[width=\linewidth]{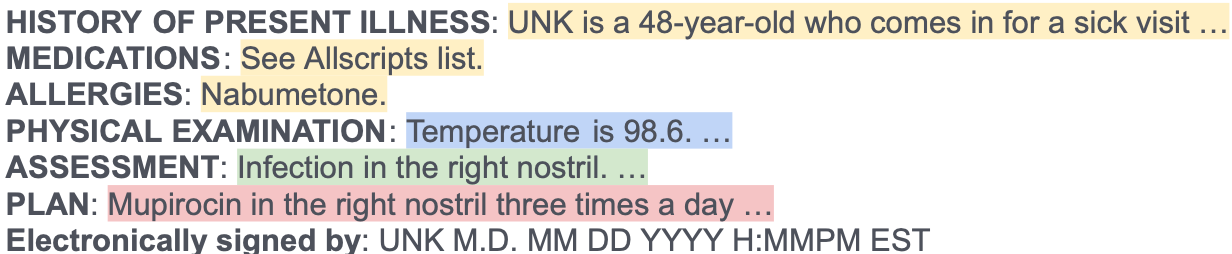}
\caption{An example with a non-explicitly structured SOAP note}
\label{fig:without_explicit}
\end{subfigure}
\vspace{1mm}
\caption{This figure represents the examples of explicitly and non-explicitly structured SOAP notes. The yellow lines, blue lines,  green lines, and redlines indicate Subjective, Objective, Assessment and Plan sections, respectively. The lines that are not colored are out-of-SOAP structure. All headers are bolded. Note that the examples are part of de-identified real clinician notes.}
\label{fig:EHR_examples}
\end{figure*}

Unfortunately, we found that EHR notes with explicit SOAP structure, when an EHR note includes all Subjective, Objective, Assessment, and Plan headers, account for only 26.37\%, leaving the majority (73.63\%) of the notes without an explicit SOAP structure. Figure~\ref{fig:EHR_examples} shows two examples of EHR notes where one with the explicit SOAP structure. Thus, supervised machine-learning approaches are nesessary to automatically identify the structure of EHR notes without explicit SOAP structure. Nonetheless, manual annotation of SOAP structure is expensive because it requisites clinical knowledge. Moreover, since each hospital has a different a SOAP note style, it is inevitable to construct a set of SOAP-labeled note data for each hospital. As a result, an innovation that does not depend or minimally depends on manual annotation data is required.

This paper introduces a novel deep learning-based automatic SOAP classification system that minimizes the dependence on manually labeled training data. For this, we first suggest a rule-based weakly SOAP labeling algorithm that utilizes 18,867 explicitly structured SOAP notes from the UMass Memorial Hospital. In addition, we present a deep learning-based SOAP classification model with a pretrained language model (LM) \cite{devlin2019bert} and bidirectional long-short term memory with conditional random field (Bi-LSTM CRF) \cite{jagannatha-yu-2016-bidirectional}. Finally, for the effective inter-hospital adaptation of the model, we propose to employ a transfer-learning framework \cite{kwon2022medjex} that reuses the parameters of the weakly trained model as initial parameters of a model instead of fine-tuning the parameters of a SOAP classification model from the scratch. 

In experiments, we compared two LMs, BioSentVec \citep{chen2019biosentvec} and BioBERT \citep{lee2020biobert}, pretrained in biomedical text corpora. To verify the effectiveness of our weakly labeling approach, we conducted experiments on four different human annotated evaluation sets collected from several hospitals and departments. The best performing system is the BioBERT-based model, achieving 89.99 on an EHR dataset from outpatient notes of the UMass Memorial Hospital. However, when we deployed the model to two other dataset from the Veterans Health Administration, the performance decreased substantially, resulting on 62.16 on an outpatients' EHR dataset and 39.28 F1 score on an EHR note dataset for urgent care patients. Moreover, we performed the models on the other publicly available dataset from the intensive care units (ICU) of the Beth Israel Deaconess Medical Center\cite{gao2022hierarchical} showing the worst performance of 13.52. Meanwhile, to demonstrate the better inter-hospital adaption of SOAP classification models with the transfer learning framework, we conducted experiments on m training data of various sizes. In all experiments, the transfer learning framework positively affected SOAP classification performance.
In particular, showing more than 20\%p performance improvement in the smallest samples of training data (50 notes), we could achieve higher performance improvement with smaller training data.


\begin{itemize}
     \item We present a framework to train machine learning classifiers with a large size of (noisily) weakly labeled data.
    \item We introduce a neural network model for SOAP classification that uses pretrained language models. 
    \item We suggest employing a transfer learning framework that utilizes the parameters of the weakly training model for a SOAP classification model's efficient inter-hospital adaptation.
    \item Experimental results show that the weakly-trained models can achieve operatable performance without hand-labeled training data. Moreover, the transfer learning framework significantly increases SOAP classification performances. 
    
\end{itemize}

\section{Related Work}

Early studies of SOAP classification used a supervised machine learning-based approach. Mowery et al, \cite{mowery2012building} represents a supervised framework for building a SOAP classification system. The authors suggest using various types of lexical, syntactic, semantic, contextual, and heuristic feature groups extracted via sophisticated feature engineering. Then a Support Vector Machine (SVM) \citep{pisner2020support} is trained with a small number of human-annotated EHR notes. Recently, deep learning-based methods using pretrained LMs are being actively studied. Schloss and Konam \citep{schloss2020towards} propose to use pretrained Embeddings from Language Models (ELMo) and a neural network model to classify a medical conversation into the SOAP sections. Nair et al.,\citep{nair2022clinical} suggest using the transfer learning method using pretrained contextual embeddings for the SOAP classification. On the other hand, since it is expensive to manually construct sufficient training data for the machine learning models, the weakly labeling approaches have been studied. Ni et al.,~\citep{ni2015fast} propose to adopt a distant supervision approach using header-to-tag map which is a knowledge-base of section headers mapping to the weak SOAP label. However, since this approach heavily relies on the pre-defined section headers, it can be problematic when a header does not belong to the knowledge-base. In particular, considering that the format of the EHR note and the headers can be different according to hospitals, departments, and even clinicians, this may limit the generalizability of the model. Note that, we also build our training dataset via weak labeling approach, but unlike the existing method, we do not rely on header information. Moreover, our SOAP classifier does not use any header information for training and inference.

\section{Automatic SOAP Classification System with the Rule-based Weakly Labeled Dataset}

\begin{figure}[!t]
\centering
    \includegraphics[width=.8\textwidth]{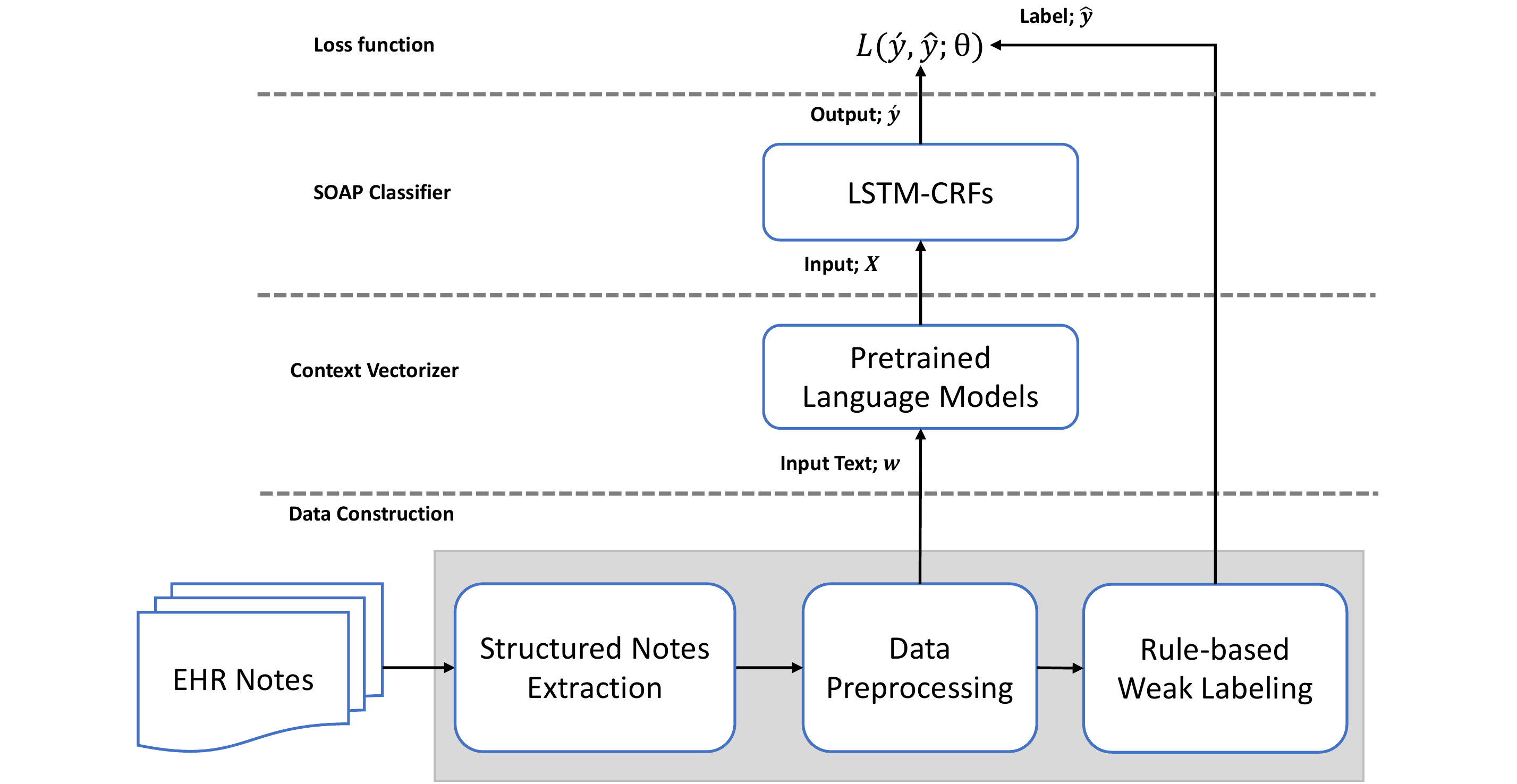}
 \caption{Illustrative concept of the automatic SOAP classification system.}
 \label{fig:amia_figure}
\end{figure}

Figure~\ref{fig:amia_figure} represents the overall structure of our automatic SOAP classification system. The system consists of the following steps: 1) Data construction, 2) Context vectorization, and 3) SOAP classifier. 

\subsection*{Task Definition}
The SOAP classification system can be defined as the \textbf{document segmentation} task and \textbf{sequence label} task. First, this system is the document segmentation task to classify documents into sentences or documents with similar topics. The input document is the patient's EHR note, and we extract the SOAP structure of paragraphs composing the note. Specifically, each paragraph should be a one of five classes which is a part of SOAP sections (`Subjective,' `Objective,' `Assessment,' and `Plan') or the out-of-SOAP, `O'. Moreover, since every paragraph in an EHR note is a one of the five labels, we can define this work as the sequence labeling task.

\subsection*{Data Construction}
This section describes how a weakly-supervision dataset for the training is made from the 71,534 de-identified EHR notes collected from the University of Massachusetts Memorial Medical Center. All patients' protected health information (PHI) was masked with a `UNK' token. The dataset consists of information on outpatients aged 18 years old or older who visited from 2014/01/01 to 2017/01/01. 




\paragraph{Structured Note Extraction}


First of all, we select \textit{explicitly structured notes} defined the EHR notes explicitly containing all of the `Subjective,' `Objective,' `Assessment,' and `Plan' headers (SOAP header). We utilize the following regular expression for the header extraction.

\begin{equation}
    \hat{~}[A-Za-z\&/:blank:]+:
\end{equation}


Herein, most notes have `Assessment' (71,534; 100\%) and `Plan' (70,260; 98.22\%) headers. On the other side, only a small portion of EHR notes explicitly include `Subjective' (20,727; 28.98\%) and `Objective' (21,340; 29.84\%) headers. Overall, we can extract 18,867 (26.37\%) notes that contains entire `Subjective,' `Objectiv,e' `Assessment,' and `Plan' headers.


\paragraph{Data Preprocessing} In data preprocessing, we first split a document into paragraphs and sentences. Then, the sequence of a single type of special characters is cleaned as a single special character. For example, a sequence of non-alphabetic or numeric characters such as `====' and `\_\_\_'. are replaced to `=' and `\_', respectively. Finally, all headers are converted to uppercase.

\begin{algorithm}[!t]
    
    \begin{algorithmic}[1]
        \caption{Pseudo code for the rule-based weak labeling}
        \label{algo:annotation}
        \Require EHR Note (\textit{D})
        
        \State $P \gets D.split(`\backslash n')$
        \State $L_0 \gets `O'$
        \State $R \gets [~]$
        \State $T \gets UTS(P)$
        \For{$P_i$, $T_i$ in $P$, $T$}
            \State $h_i \gets get\_header(P_i)$
            \If{$h_i$ in $H$}
                \State $L_i \gets h_i$
            \ElsIf{$L_{i-1}$ is `Plan' and $T_{i-1} != T_i$}
                \State $L_i \gets `O'$
            \EndIf
            \State $R \gets R + [(P_i, L)]$
        \EndFor

        \State \textbf{Return} $R$
        
    \end{algorithmic}
    
\end{algorithm}
\paragraph{Rule-based Weak Labeling} Although we choose the structured notes, paragraphs without the SOAP headers also can be a part of the SOAP structure. For example, paragraphs started with headers such as \textit{history} and \textit{current medication} can be considered as subjective sections, while \textit{examination} can be regarded as objective setion. Moreover, in many cases, such as when the previous paragraph and topic are the same, there may not be an explicit header.  Thus, if it is regarded that only paragraphs including the SOAP header are part of the SOAP structure in the training dataset, many patterns may be missed. To ameliorate this issue, we suggest a simple labeling rule the dataset with the following assumptions:
\begin{enumerate}
    \item In the EHR note, the SOAP structure should be in the Subjective-Objective-Assessment-Plan order
    \item If the header of a paragraph is part of the SOAP header set that consists of Subjective, Objective, Assessment, Plan, or the predefined header set introduced by Mowery et al \cite{mowery2012building} the paragraph is annotated as the corresponding label
    \item Paragraphs between paragraphs in which the SOAP header has the same label with the previous paragraph. That is, all paragraphs placed after the paragraph with the Subjective header and before the paragraph with the Objective header are annotated as Subjective
\end{enumerate}

Algorithm~\ref{algo:annotation} is a pseudo code for the annotation rule that reflects our assumptions where $D$ is an input EHR note and $H$ the SOAP header set. To begin with, an input EHR note $D$ is split as a sequence of lines (line 1). Then, we initialize the current label variable $L$ as `O' that indicates the out-of-SOAP section. In line 3, $R$ is the return value and it is set to a blank list. We perform topic segmentation, clustering the text of the document according to the topic, and store the result in $T$ (line 4). As a result, $T$ represents whether each paragraph has the same topic as the previous paragraph in binary encoding form \citep{purver2011topic}. Herein, we utilize the unsupervised text segmentation (UTS) tool \citep{Wanf2016uts} for topic segmentation. Lines 5-13 are the process of defining the label of each paragraph. In line 6, we explicit the header of $i^{th}$ paragraph $P_i$ by using the aforementioned regular expression. If the header includes in $H$ (line 7), then $L$ is set to the $h_i$. Otherwise, if $h_i$ is not the SOAP header but $L$ is `Plan' and the topic of the current paragraph ($T_i$) is different from the topic of the previous paragraph ($T_{i-1}$), then we set $L$ as `O' (Lines 9-11). This is because text not related to SOAP structure such as closing and clinician's signature can place at the end of the EHR note. Subsequently, $R$ is updated by appending the pair of the current paragraph $P_i$ and $L$.

\subsection*{Context Vectorization}
In the context vectorizer, input text ($\mathbf{w}$) is transformed into a feature vector that is an input of machine learning models. Traditionally, in the BioNLP domain, lexical features (bag-of-words, word n-grams ...), grammatical features (part-of-speech tags, parsing tree, ...), and semantic features (semantic type, named-entity types ...) used for input features \citep{mowery2012building, athenikos2010biomedical}. After that, with the emergence of deep learning, many studies utilized word distributed word vector features \citep{moen2013distributional, lee2016ksanswer}. Recently, contextualized word vector representation that creates context-sensitive word vectors \citep{ethayarajh2019contextual} by using pretrained LMs are in the spotlight. Especially, since the advent of BERT \citep{devlin2019bert}, contextualized word vector representations overwhelm previous approaches in most NLP domains \citep{rothman2021transformers}. In this paper, by following cutting-edge technological trends, we extract input feature vectors with pretrained LMs. The paragraphs belonging to the input text $\mathbf{w}$ are input to the pretrained LMs and converted into the fixed-length vector. Finally, the paragraph vectors are concatenated to form the input matrix $\mathbf{X}$. 

In this work, for the context vectorization, BioSentVec \citep{chen2019biosentvec} and BioBERT \citep{lee2020biobert} models are used as pretrained LM designed to be specialized in the Biomedical text domain. Herein, we describe how can we create the paragraph vector from an input paragraph via BioSentVec and BioBERT.

\paragraph*{BioSentVec} We first split an input paragraph into the sequence of words with the model's tokenizer. The sequence of tokenized words is transferred to the sequence of word vectors via pretrained word embedding. Then, the word vectors are input to the model generating paragraph vector.

\paragraph*{BioBERT} Since BERT models have a strict limitation in the input token length of 512, some long paragraphs are difficult to handle with the model. Thus, an input paragraph is split as a sequence of sentences with a sentence splitter. Then, a sentence are tokenized into the sequence of wordpiece tokens \citep{devlin2019bert} with the model's tokenizer. The sequence of tokens is input to the model then we can get hidden vector representations of the tokens. Then, mean pooling is utilized to generate the sentence vector of the input sentence. Finally, the paragraph vector is created by mean pooling of the vector representations of the sentences that belong to the paragraph. 

\subsection*{SOAP Classifier}
In this study, the SOAP classifier uses Bi-LSTM-CRFs, which has been widely used in biomedical named entity recognition tasks \citep{jagannatha-yu-2016-bidirectional}\citep{jagannatha-yu-2016-structured}. The model comprises two layers 1) the Bi-LSTM layer and 2) the CRF layer. 

The Bi-LSTM layer is a stacked Bi-LSTM that is comprised of multi-layer Bi-LSTM. Herein, in Eq~\ref{equ:bi-lstm}, $\mathbf{X}$ is input to Bi-LSTM layer resulting hidden $\mathbf{H}$. Then, logit $\mathbf{S}$ can be calculated via linear transformation of $\mathbf{H}$ Eq~\ref{equ:linear}. In this case, each dimension of $\mathbf{S}$ represents the observation scores of the output labels.

\vspace{-3mm}
\begin{equation}
    \label{equ:bi-lstm}
    \mathbf{H} = Bi-LSTM(\mathbf{X})
\end{equation}
\vspace{-1mm}
\begin{equation}
    \label{equ:linear}
    \mathbf{S} = Linear(\mathbf{H})
\end{equation}

Finally, the observation score is input to the conditional random field (CRF) \citep{lafferty2001conditional} layer. Herein, let's assume that $y$ is a sequence of output labels. Then, we can calculate the score of the sequence $y$ defined as Eq.~\ref{equ:crfs} with the transition matrix $T$. Then, in Eq.~\ref{equ:crfs}, we jointly decode the optimal output sequence $\hat{y}$ from all possible sequences of labels $\mathbf{Y}$.

\vspace{-3mm}
\begin{equation}
    \label{equ:crfs}
    f(y)=exp\left( \sum_{i=0}^{n}\mathbf{T}_{y_iy_{i+1}}+\sum_{i=0}^{n}{\mathbf{S}_{iy_i}}\right)
\end{equation}
\vspace{-1mm}
\begin{equation}
    \label{equ:optimal}
    \hat{y}=\underset{\tilde{y} \in \bf{Y}}{\argmax}~f(\tilde{y})
\end{equation}

\section{Transfer-learning Framework for Efficient Inter-hospital Adaptation of a Model}

Although the SOAP structure helps clinicians to write progress notes structurally, since the required information to be described is different for each hospital, department, and patient, the structure of the SOAP note may vary \cite{} which makes inter-hospital adaptation of the models more challenging. Therefore, it is difficult to avoid constructing manually-annotated SOAP datasets for each hospital or even department in a hospital. However, the annotators of SOAP classification should have a clinical background. Tremendous costs are required to construct such a large size SOAP annotated dataset for every hospital.

To ameliorate the aforementioned issue, we suggest employing the transfer learning framework \citep{kwon2022medjex} that fine-tunes SOAP classifiers with trained parameters rather than with randomly initialized parameters. Specifically, we first make a weakly trained model with our rule-based labeled data then re-train the parameters of the weakly trained model with SOAP annotated notes from another hospital. Through this, it is expected that information necessary for inter-hospital migration, such as SOAP structure and similar header information, can be used in the weakly trained model.

\section{Experiments}
\subsection*{Experimental Dataset}

In this work, we prepared five different types of the dataset collected from three different EHR databases: University of Massachusetts Memorial Medical Center (UMass$_{Weak}$, UMass$_{Outpatient}$), Veterans Affairs (VA$_{Outpatient}$, VA$_{UC}$) and Beth Israel Deaconess Medical Center (MIMIC). A detailed description of each dataset is given below.

The first dataset is UMass$_{Weak}$ includes EHR notes labeled by our weakly-labeling algorithm. Herein, the UMass$_{Weak}$ is randomly split into a 15,093 train set (80\%), 1,887 validation set (10\%), and 1,887 test set (10\%). Using this dataset, we train the weakly-trained SOAP classifiers and evaluate the vector representation extracted from pretrained LMs.

MIMIC \citep{gao2022hierarchical} is a publicly available SOAP classification dataset that is a part of the progress note understanding tasks. The dataset is made up of 678 human-annotated intensive care notes from the MIMIC database \citep{johnson2016mimic}. The dataset is split into 586 notes for the train set, 75 notes for the validation set, and 87 notes for the test set. Note that, in this dataset, the `Assessment' and `Plan' sections are not separated but regarded as one section named `Assessment \& Plan.' The dataset is available through PhysioNet \citep{Gao2022-ep}.

On the other hand, we prepare additional two evaluation datasets which are annotated by human annotators. The first dataset is made up of 50 outpatient EHR notes collected from the UMass Medical School database (UMass$_{Outpatient}$). The other dataset is composed of 50 EHR notes collected from Veterans Affairs (VA) Corporate Data Warehouse (CDW) and Text Integration Utilities (TIU). We randomly selected 25 primary care outpatient notes (VA$_{Outpatient}$) which is similar to UMass$_{Outpatient}$ and 25 primary care urgent care notes as a comparison (VA$_{UC}$). Since doctors from the same hospital may use the same built-in SOAP notes templates to write notes \citep{vadocument, rogers_2021}, we collected each note from a different VA hospital site to test the universality of our algorithm. 
Two annotators separately annotates each EHR note. Then, we calculate inter-annotator agreement to check the reliability of the manual annotations via Cohen's Kappa \citep{artstein2008inter}. As a result, the inter-annotator agreement is 90.67 (almost perfect agreement). We randomly select one of the annotator's answers when the labels disagree. 

We have the three main questions we would like to address with the dataset: 

\begin{enumerate}
    \item Do weakly-trained models predict SOAP structures of EHR notes from the same hospital? (UMass$_{Outpatient}$)
    \item Do weakly-trained models predict SOAP structures of EHR notes from other hospitals? (VA$_{Outpatient}$, VA$_{UC}$, and MIMIC)
    \item Does the transfer learning framework help to improve the performance in another hospital? (MIMIC)
\end{enumerate}

   


\subsection*{Experimental Setting}
In preprocessing, we utilized medspaCy \citep{eyre2021medspacy} as a text analyzer. All pretrained LMs in the context vectorizer are set to default. For training experimental models, we set hyper-parameters as follows. 
Batch size, learning rate, and maximum epoch were set to 64, 5e-3 and 10. In addition, the Bi-LSTM layer is composed of stacked Bi-LSTM of 3 layers and the size of hidden nodes is 128. During model training, cross-entropy loss \citep{huang2015bidirectional} is utilized and only the parameters for the SOAP classifier are updated while the parameters of the pretrained LMs are frozen.

All experiments were conducted in Ubuntu 20.04 environment using one RTX6000 GPU, Intel Xeon Gold 6226R CPU, and 755GB RAM. In addition, we conducted the t-test \citep{yang1999re} to verify statistical significance on the performance difference between experimental results. Furthermore, the F1 score, the harmonic mean of precision of Eq.~\ref{equ:prec}, and recall of Eq.~\ref{equ:rec}, is used for performance evaluation criteria.

\vspace{-3mm}
\begin{equation}
\small
\label{equ:prec}
Precision = \frac{\#~of~true~positive~SOAP~labeled~paragraphs}{\#~of~positive~SOAP~labeled~paragraphs}
\end{equation}

\begin{equation}
\small
\label{equ:rec}
Recall = \frac{\#~of~true~positive~SOAP~labeled~paragraphs}{\#~of~true~SOAP~labeled~paragraphs}
\end{equation}

\subsection*{Experimental Results}
Table~\ref{tab:experimental_results_umass_weak} is the experimental results of weakly trained models on evaluation sets. Herein, the evaluation sets consist of and UMass$_{Outpatient}$, VA$_{Outpatient}$ and VA$_{UC}$ and the MIMIC test set. 


\begin{table}[b]
\centering
\small
\begin{tabular}{c|c|c|c|c|c|c|c}
\hline
Evaluation Set & LM & S & O & A & P & Out & Macro Avg. \\
\hline\hline
UMass$_{Outpatient}$ & \multirow{4}{*}{BioSentVec} & 86.15 & 96.92 & 80.30 & 88.68 & 90.01 & 88.41\\
\cline{3-8}
VA$_{Outptient}$ & & 53.18 & 77.47 & 68.85 & 81.82 & 34.52 & \textbf{63.17} \\
\cline{3-8}
VA$_{UC}$ & & 48.85 & 77.23 & 52.46 & 31.58 & 1.12 & \textbf{42.25}\\
\cline{3-8}
MIMIC &  & 53.11 &  0.00 & \multicolumn{2}{c|}{18.71} & 0.00 & \textbf{17.95}\\
\hline
UMass$_{Outpatient}$ &  \multirow{4}{*}{BioBERT} & 89.66 & 92.91 & 84.30 & 88.89 & 94.19 & \textbf{89.99} \\
\cline{3-8}
VA$_{Outptient}$      & &  63.27 & 84.42 & 62.50 & 61.9 & 38.71 & 62.16 \\
\cline{3-8}
VA$_{UC}$ & & 30.15 & 77.73 & 70.59 & 16.22 & 1.72 & 39.28 \\
\cline{3-8}
MIMIC & &45.94 &5.84  & \multicolumn{2}{c|}{2.33} &  0.00 & 13.52 \\
\hline
\end{tabular}
\caption{Experimental results on the zero-shot evaluation sets (UMass$_{Weak}$, UMass$_{Outpatient}$, VA$_{Outptient}$, VA$_{UC}$, and MIMIC). Herein, `S', `O',`A', `P' and `Out' means `Subjective', `Objective',`Assessment', `Plan', and `Out-of-SOAP', respectively. Note that, since the assessment and plan sections are not separated but annotated as `Assessment and Plan' in MIMIC, the performance of `A' and `P' sections are merged. Macro Avg. indicates the average F1 score of the classes.}
\label{tab:experimental_results_umass_weak}
\end{table}




\paragraph{UMass$_{Outpatient}$}  In UMass$_{Outpatient}$, the BioBERT-based model shows the performance of 89.99 which denotes that the model trained with our weakly labeled data can be useful in real data. In addition, the performance of BioBERT-based model is 1.58\%p higher than that of BioSentVec-based model. However, the overall performance of the weakly labeled test set is lower than that of the weakly labeled test set. In particular, the performance is markedly decreased in the Subjective and Assessment sections. First of all, in the case of Subjective, there are many cases of misclassifying the Chief Complaint (CC) section as the `Out' section. This is because, in some cases, the CC section locates before the explicit Subjective section. Therefore, some CC sections are annotated as `Out' with our rule-based weak labeling, acting as noise in model training. Otherwise, in the Assessment section, most error cases occur in psychiatric notes. Especially, the SOAP classifiers frequently fail to analyze the multi-axial system-based diagnosis \citep{goldman1992revising}.

\paragraph{The VA dataset} By comparing UMass$_{Outpatient}$, we can observe substantial performance decreases in VA$_{Outpatient}$. One potential reason is that the UMass notes include more free-text sentences, while VA$_{Outpatient}$ tends to be structured as semi-structured notes. For example, only 2 (out of 50) Assessments of UMass outpatient notes were written in list format, while 12 (out of 25) Assessments of VA outpatient notes were written in list format. On the other hand, the performance dramatically dropped in VA$_{UC}$, because the gap with the training data of outpatients is large due to the characteristics of urgent care being different. In addition, some elements of the SOAP structure may be omitted. For example, all UMass$_{Outpatient}$ notes include the Assessment and Plan sections. However, in the 25 VA$_{UC}$ notes, only 11 have the Assessment section and only 5 have the Plan section. Finally, the BioSentVec-based model shows higher performance than the BioBERT-based model which means that it is advantageous for generalization.

We found that these performance decrement in the `Out' section could be accounted for the discrepancies of the header formats of the weakly labeled data and the VA dataset. For example, the headers of the out section in the UMass dataset usually contains `Electronically signed by', 'Attending'. Otherwise, the headers of the out section in the VA dataset `D (Date)', `T (Time)' which are differently presented from the that of the UMass dataset. 

\paragraph{MIMIC} Both BioSent2Vec-based and BioBERT-based models show the lowest performance in MIMIC test set. Herein, BioBERT-based model has significantly lower performance of 13.52 compared to the that of BioSentVec-based model's 17.95. Especially, both models could not identify `out' sections at all and the performance dramatically decreased in the objective section. From the results, we found that not only the objective section header format is different from the weakly-labeled data, but also the weakly-trained model failed to properly detect it because there are many laboratory tests due to the characteristics of ICU notes. In the same way, the header format of `out' sections were also divergent from the that of the weakly-labeled dataset including `Communication', `Code status', and `Disposition'.


\subsection*{Experimental Results on Transfer Learning} 

To verify that the transfer learning framework is advantageous for the models' inter-hospital adaptation, we trained SOAP classification models on the MIMIC train data with the following two environments. The first setting is `Finetune' where the parameters of the SOAP classifier of a model were randomly initialized and fine-tuned with the trained set of MIMIC dataset. The other one is `Transfer' where the parameters of the SOAP classifier of a model were initilized with the pre-trained parameters of the UMass$_{Weak}$ model. Moreover, we trained models with the different sizes of data randomly sampled from the MIMIC training data. Finally, to minimize the effect of random sampling on data, we trained the models by three times on different random seeds for each environment and data size, and calculated the mean and variance.  

Table~\ref{tab:transfer_learning} demonstrates the experimental results on the transfer learning framework. We can see that the models trained in the transfer environment always have higher performance compared to those of the finetune environment achieving significantly higher performance ($p<0.05$) in eight out of ten settings. Especially, we could get the higher the performance improvement as the smaller the training data size. The BioSentVec-based model trained with 50 notes got the highest performance enhancement of 28.89. The results supporting our assumption that a weakly trained model can help to reach higher SOAP classification performance with less training data via the transfer learning framework. Moreover, in nine of the ten settings, the standard deviations were smaller in the transfer environment. It indicates that the models tend to be trained more stably in a transfer environment.

\begin{table}[]
\centering
\begin{tabular}{c|c|cc|cc|c|c}
\hline
\multirow{2}{*}{Model} & \multirow{2}{*}{Train Size} & \multicolumn{2}{c|}{Finetune} & \multicolumn{2}{c|}{Transfer} & \multirow{2}{*}{$\Delta$} & \multirow{2}{*}{P-value} \\
\cline{3-6}
 & & Avg. & Std. & Avg. & Std. & & \\
\hline\hline
\multirow{5}{*}{BioSentVec} & 50 & 56.16 & 1.53 & 85.05 & 3.62 & \textbf{28.89} & 0.005 \\
 & 100 & 66.35 & 6.46 & 89.23 & 1.86 & \textbf{22.88} & 0.018 \\
 & 200 & 84.80 & 2.90 & 90.45 & 0.27 & 5.65 & 0.044 \\
 & 400 & 87.70 & 2.07 & 91.72 & 0.88 & \textbf{4.02} & 0.015 \\
 & Full & 92.05 & 1.15 & 93.12& 0.59 &\textbf{ 1.07} & 0.099 \\
\hline
\multirow{5}{*}{BioBERT} & 50 & \textbf{66.37} & 3.16 & \textbf{90.54} & 0.85 & 24.18 & 0.003 \\
 & 100 & \textbf{71.52} & 1.39 & \textbf{91.56} & 0.63 & 20.04 & 0.002 \\
 & 200 & \textbf{85.71} & 4.36 & \textbf{92.88} & 0.25 & \textbf{7.17} & 0.048 \\
 & 400 & \textbf{89.79}& 2.87 & \textbf{93.39} & 0.44 & 3.60 & 0.082 \\
 & Full & \textbf{92.98} & 0.05 & \textbf{93.81} & 0.28 & 0.83 & 0.018 \\
\hline

\end{tabular}
\caption{Experimental results on the transfer-learning framework trained on different sizes of MIMIC training dataset. Herein, `Train Size' means the number notes randomly sampled from MIMIC training dataset. We trained the models with three different random seeds reporting the average (Avg.) of Macro F1s, the standard deviation (Std.). Furthermore, `Finetune' is when the parameters in SOAP classifiers are randomly initialized. Meanwhile, `Transfer' is when the parameters in SOAP classifiers are initialized with parameters pretrained on UMass$_{Weak}$ train data. Finally, `$\Delta$' means the difference between the average values on the finetune and transfer settings finetune and transfer settings. `P-value' is the statistical significance of the difference.}
\label{tab:transfer_learning}
\end{table}

\begin{table}[]
\centering
\begin{tabular}{c|l|c|cccc}
\hline
    & Train Size & UMass$_{Weak}$ & UMass$_{Outpatient}$ & VA$_{Outpatient}$ & VA$_{UC}$  & MIMIC  \\
\hline\hline
\multirow{6}{*}{BioSentVec} & 100   & 30.23  & 33.82  & 20.16  & \textbf{18.00} & \textbf{17.56} \\
    & 500   & 91.00  & 80.40  & 49.05  & \textbf{31.14} & \textbf{18.53} \\
    & 1,000   & 94.57  & 82.50  & 53.77  & \textbf{30.98} & \textbf{19.70} \\
    & 5,000   & 97.67  & 87.38  & 58.06  & \textbf{37.67} & \textbf{18.69} \\
    & 10,000   & 97.91  & 88.02  & 57.23  & \textbf{36.36} & \textbf{14.83} \\
    & 15,903   & 98.15  & 88.41  & \textbf{63.17} & \textbf{42.25} & \textbf{17.95} \\
    \hline
    \multicolumn{2}{c|}{Spearman's $\rho$}& 0.94 & 1.00 & 0.94 & 0.89 & -0.14 \\
\hline
\multirow{6}{*}{BioBERT} & 100   & \textbf{38.86} & \textbf{47.96} & \textbf{29.77} & 9.13  & 15.56  \\
    & 500   & \textbf{93.70} & \textbf{83.73} & \textbf{58.39} & 26.28  & 13.52  \\
    & 1,000   & \textbf{94.91} & \textbf{85.52} & \textbf{58.83} & 22.30  & 11.89  \\
    & 5,000   & \textbf{97.89} & \textbf{89.20} & \textbf{64.80} & 38.14  & 12.50  \\
    & 10,000   & \textbf{97.97} & \textbf{88.45} & \textbf{62.42} & 33.09  & 13.76  \\
    & 15,903   & \textbf{98.24} & \textbf{89.99} & 62.16  & 39.28  & 13.52 \\
    \hline
    \multicolumn{2}{c|}{Spearman's $\rho$} & 1.00 & 0.94 & 0.77 & 0.89 & -0.32\\
\hline
\end{tabular}
\caption{Macro F1s of weakly trained models on different sizes of randomly sampled UMass$_{Weak}$ train data. }
\label{tab:size_of_weakly_label}
\end{table}

\section*{Discussion}
\subsection*{How Does the Size of Weakly Labeled Train Data Impacts on Evaluation Sets?}
Although the weakly labeling on EHR notes is beneficial to SOAP classifiers without additional annotations, it is still questionable whether a large amount of EHR notes is required. Thus, we conducted additional experiments on the impact of the size of the weakly labeled notes. To be specific, SOAP classifiers were trained with the difference sizes (100, 500, 1,000, 5,000, 10,000 and 15,903) of randomly sampled weakly labeled training set, then the performance of the classifiers were assessed on the evaluation sets. Herein, we calculated the Spearman's rank correlation coefficient (Separman's $\rho$) \citep{dancey2007statistics} to analyze the correlation between the size of data and the performance. 

Table~\ref{tab:size_of_weakly_label} demonstrates the experimental results of the impact on the size of the weakly labeled notes. The results shows that as the size of the training set increases, the performance tends to improve without MIMIC. In UMass$_{Weak}$, the performances were monotonically increased as the size of training set indicating that the models could accurately recognize the weakly labeling algorithm's pattern with larger training samples. Meanwhile, the performance in UMass$_{Outpatient}$ enhanced more than 40\%p in both of BioSentVec and BioBERT based models showing very strongly monotonic increasing (both $\rho>0.70$). Moreover, the performance increased more than twice in both VA evaluation sets presenting very strongly ($\rho>0.70$) positive correlationship. However, in MIMIC, the performances have negligibly ($0.00>\rho>-0.20$) or moderate ($-0.30>\rho>-0.40$) negative correlation coefficient. As a result, a larger size of weakly labeling training data leads performance increment in target (UMass$_{Outpatient}$) or relevant (VA$_{Outpatient}$ and VA$_{UC}$) domain. Otherwise, the size of a weakly labeled train set does not benefit to the performance. Additional analyses on the lexical and header overlappings can be found in the supplementary materials.

\subsection*{Execution Time Comparison} We gauge the running time of the entire 18,867 weakly labeled dataset. Table~\ref{tab:time_comparison} demonstrates the results of the execution time comparison of the BioSentVec and the BioBERT based models. Note that data preprocessing has the same execution time for both models since it is a common process between the models. Similarly, in SOAP classifier, paragraph vectors of the same length are input in both models, and the only difference is the input vector dimension, so the difference in execution time is trivial. 
Otherwise, the experimental results show that BioSentVec is overwhelmingly faster because of the discrepancies in the models' complexity. Thus, since the performance of the BioBERT-based model is higher, we should consider the tradeoff between the running time and the performance when we choose an appropriate model for downstream tasks. 
\begin{table}[]
\centering
\begin{tabular}{c|c|c}
\hline
\multicolumn{1}{c|}{}                & Model       & Execution Time \\
\hline\hline
Data Preprocessing                       & -           & 1h 4m 29s      \\
\hline
\multirow{2}{*}{Context Vectorizer} & BioSentVec & \textbf{1m 36s}    \\
\cline{2-3}
                                    & BioBERT     & \textbf{20m 37s}     \\
\hline                                
\multirow{2}{*}{SOAP Classifier} & BioSentVec & 2m 48s       \\
\cline{2-3}
                                    & BioBERT     & 2m 50s    \\
\hline
\end{tabular}
\caption{Experimental results on the execution time comparison.}
\label{tab:time_comparison}
\end{table}

\subsection*{Error Analysis}


\begin{figure}[!t]
\centering
\includegraphics[width=.7\textwidth]{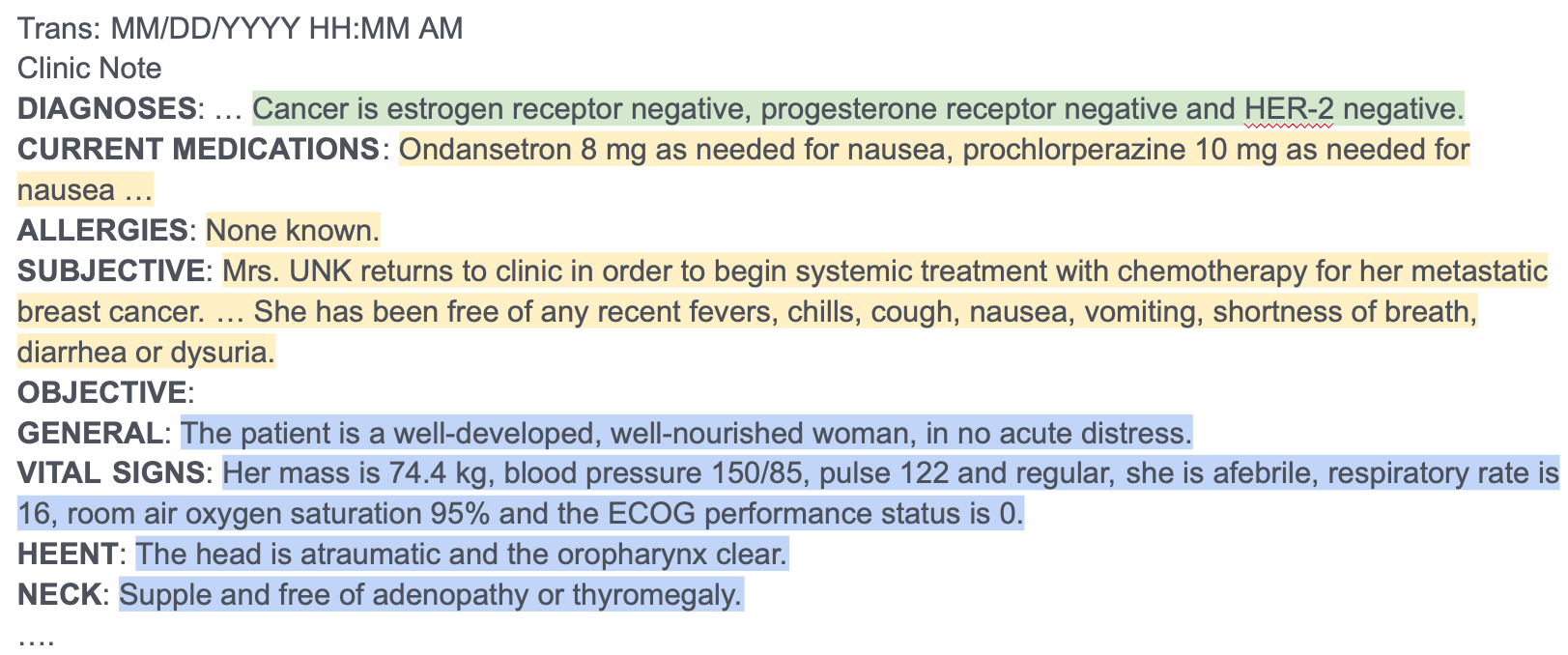}
 \caption{An example of an error case when the order of SOAP structure is not typical. The green-colored line indicates the Assessment section. Yellow-colored lines are the Subjective section. The blue-colored lines denote the Objective Section. }
 
 \label{fig:error_case}
\end{figure}


One representative error type is observed when the order of the SOAP section is not the typical Subjective-Objective-Assessment-Plan order. Figure~\ref{fig:error_case} is an example de-identified EHR note in UMass$_{Outpatient}$. Herein, the green-colored Assessment section is placed before the Subjective. In this case, the BioBERT-based model predicts the Assessment section as Subjective showing that our model not only uses text information but also depends on the typical order of the SOAP structure. Indeed, when the paragraph is located after the objective sections, the model appropriately classifies the paragraph as the Assessment.

\subsection*{Limitation}
Thorough the experimental results, we can notice that the weakly trained models successfully predicted the SOAP structures in the target domain, while the models failed in inter-hospital adaptation. This is due to the SOAP template used differently in each hospital and department, and in particular, the weakly trained models suffered from unseen headers. Indeed, header features are known as major features in SOAP classification \citep{mowery2012building}. We also verified the importance of header features through the experiments included in the supplementary material.


\section{Conclusion and Future Work}
This paper introduces weakly supervision approach . Moreover, we suggest to employ the transfer framework that utilize the pretrained parameters of the weakly trained models for the data efficient the SOAP classification training. We evaluated the SOAP classifier on various hand-labeled dataset. Experimental results show that our model can properly perform the SOAP classification in EHR notes gathered from the target hospital. In addition, although the performances on the EHR notes extracted from other hospitals and departments were decreased, we found that the transfer learning framework led better results with a small size of expert-annotated dataset.

In the future, we plan to explore advanced methodologies to achieve better results on the zero-shot setting. Especially, prompt-based approaches \citep{sun2021nsp, liu2021pre} that concatenate additional text that can represent the target task into an original input text. In addition, automatic local analysis approaches (e.g. semi-supervision \citep{kim2017method}, pseudo relevance feedback \citep{gurulingappa2016semi} ...) can be employed to increase the variance of the predefined SOAP header set.

\section*{Acknowledgments}
This study was performed in accordance with the recommendations laid out in the World Medical Association Declaration of Helsinki. The study protocol was approved by the Institutional Review Boards (IRBs) of University of Massachusetts Medical School (IRB No. H00008467) and Veterans Health Administration (IRB No. 1652850).

\makeatletter
\renewcommand{\@biblabel}[1]{\hfill #1.}
\makeatother

\bibliographystyle{vancouver}
\bibliography{jamia}  

\begin{thebibliography}{10}

\bibitem{jagannatha2019overview}
Jagannatha A, Liu F, Liu W, Yu H.
\newblock Overview of the first natural language processing challenge for
  extracting medication, indication, and adverse drug events from electronic
  health record notes (MADE 1.0).
\newblock Drug safety. 2019;42(1):99-111.

\bibitem{mitra2021risk}
Mitra A, Ahsan H, Li W, Liu W, Kerns RD, Tsai J, et~al.
\newblock Risk Factors Associated With Nonfatal Opioid Overdose Leading to
  Intensive Care Unit Admission: A Cross-sectional Study.
\newblock JMIR medical informatics. 2021;9(11):e32851.

\bibitem{liu2018deep}
Liu J, Zhang Z, Razavian N.
\newblock Deep ehr: Chronic disease prediction using medical notes.
\newblock In: Machine Learning for Healthcare Conference. PMLR; 2018. p.
  440-64.

\bibitem{zhang2020combining}
Zhang D, Yin C, Zeng J, Yuan X, Zhang P.
\newblock Combining structured and unstructured data for predictive models: a
  deep learning approach.
\newblock BMC medical informatics and decision making. 2020;20(1):1-11.

\bibitem{weed2014medical}
Weed LL.
\newblock Medical records that guide and teach.
\newblock Clinical Problem Lists in the Electronic Health Record. 2014;19.

\bibitem{yang2020generating}
Yang Z, Yu H.
\newblock Generating Accurate Electronic Health Assessment from Medical Graph.
\newblock In: Proceedings of the Conference on Empirical Methods in Natural
  Language Processing. Conference on Empirical Methods in Natural Language
  Processing. vol. 2020. NIH Public Access; 2020. p. 3764.

\bibitem{gao2022hierarchical}
Gao Y, Dligach D, Miller T, Tesch S, Laffin R, Churpek MM, et~al.
\newblock Hierarchical Annotation for Building A Suite of Clinical Natural
  Language Processing Tasks: Progress Note Understanding.
\newblock In: Prceedings of LREC; 2022. .

\bibitem{devlin2019bert}
Devlin J, Chang MW, Lee K, Toutanova K.
\newblock BERT: Pre-training of Deep Bidirectional Transformers for Language
  Understanding.
\newblock In: Proceedings of NAACL; 2019. p. 4171-86.

\bibitem{jagannatha-yu-2016-bidirectional}
Jagannatha AN, Yu H.
\newblock Bidirectional {RNN} for Medical Event Detection in Electronic Health
  Records.
\newblock In: Proceedings of the 2016 Conference of the North {A}merican
  Chapter of the Association for Computational Linguistics: Human Language
  Technologies. San Diego, California: Association for Computational
  Linguistics; 2016. p. 473-82.
\newblock Available from: \url{https://aclanthology.org/N16-1056}.

\bibitem{kwon2022medjex}
Kwon S, Yao Z, Jordan HS, Levy DA, Corner B, Yu H.
\newblock MedJEx: A Medical Jargon Extraction Model with Wiki's Hyperlink Span
  and Contextualized Masked Language Model Score.
\newblock In: Proceedings on the 2022 Empirical Methods in Natural Language
  Processing; 2022. .

\bibitem{chen2019biosentvec}
Chen Q, Peng Y, Lu Z.
\newblock BioSentVec: creating sentence embeddings for biomedical texts.
\newblock In: 2019 IEEE International Conference on Healthcare Informatics
  (ICHI). IEEE; 2019. p. 1-5.

\bibitem{lee2020biobert}
Lee J, Yoon W, Kim S, Kim D, Kim S, So CH, et~al.
\newblock BioBERT: a pre-trained biomedical language representation model for
  biomedical text mining.
\newblock Bioinformatics. 2020;36(4):1234-40.

\bibitem{mowery2012building}
Mowery D, Wiebe J, Visweswaran S, Harkema H, Chapman WW.
\newblock Building an automated SOAP classifier for emergency department
  reports.
\newblock Journal of biomedical informatics. 2012;45(1):71-81.

\bibitem{pisner2020support}
Pisner DA, Schnyer DM.
\newblock Support vector machine.
\newblock In: Machine learning. Elsevier; 2020. p. 101-21.

\bibitem{schloss2020towards}
Schloss B, Konam S.
\newblock Towards an automated SOAP note: classifying utterances from medical
  conversations.
\newblock In: Machine Learning for Healthcare Conference. PMLR; 2020. p.
  610-31.

\bibitem{nair2022clinical}
Nair N, Narayanan S, Achan P, Soman K.
\newblock Clinical Note Section Identification Using Transfer Learning.
\newblock In: Proceedings of Sixth International Congress on Information and
  Communication Technology. Springer; 2022. p. 533-42.

\bibitem{ni2015fast}
Ni J, Delaney B, Florian R.
\newblock Fast Model Adaptation for Automated Section Classification in
  Electronic Medical Records.
\newblock Studies in health technology and informatics. 2015;216:35-9.

\bibitem{purver2011topic}
Purver M.
\newblock Topic segmentation.
\newblock Spoken language understanding: systems for extracting semantic
  information from speech. 2011:291-317.

\bibitem{Wanf2016uts}
Liang W. Unsupervised Text Segmentation. GitHub; 2016.
\newblock \url{https://github.com/intfloat/uts}.

\bibitem{athenikos2010biomedical}
Athenikos SJ, Han H.
\newblock Biomedical question answering: A survey.
\newblock Computer methods and programs in biomedicine. 2010;99(1):1-24.

\bibitem{moen2013distributional}
Moen S, Ananiadou TSS.
\newblock Distributional semantics resources for biomedical text processing.
\newblock Proceedings of LBM. 2013:39-44.

\bibitem{lee2016ksanswer}
Lee Hg, Kim M, Kim H, Kim J, Kwon S, Seo J, et~al.
\newblock KSAnswer: Question-answering system of kangwon national university
  and sogang university in the 2016 BioASQ challenge.
\newblock In: Proceedings of the Fourth BioASQ workshop; 2016. p. 45-9.

\bibitem{ethayarajh2019contextual}
Ethayarajh K.
\newblock How Contextual are Contextualized Word Representations? Comparing the
  Geometry of BERT, ELMo, and GPT-2 Embeddings.
\newblock In: Proceedings of the 2019 Conference on Empirical Methods in
  Natural Language Processing and the 9th International Joint Conference on
  Natural Language Processing (EMNLP-IJCNLP); 2019. p. 55-65.

\bibitem{rothman2021transformers}
Rothman D.
\newblock Transformers for Natural Language Processing: Build innovative deep
  neural network architectures for NLP with Python, PyTorch, TensorFlow, BERT,
  RoBERTa, and more.
\newblock Packt Publishing Ltd; 2021.

\bibitem{jagannatha-yu-2016-structured}
Jagannatha A, Yu H.
\newblock Structured prediction models for {RNN} based sequence labeling in
  clinical text.
\newblock In: Proceedings of the 2016 Conference on Empirical Methods in
  Natural Language Processing. Austin, Texas: Association for Computational
  Linguistics; 2016. p. 856-65.
\newblock Available from: \url{https://aclanthology.org/D16-1082}.

\bibitem{lafferty2001conditional}
Lafferty JD, McCallum A, Pereira FC.
\newblock Conditional Random Fields: Probabilistic Models for Segmenting and
  Labeling Sequence Data.
\newblock In: Proceedings of the Eighteenth International Conference on Machine
  Learning; 2001. p. 282-9.

\bibitem{johnson2016mimic}
Johnson AE, Pollard TJ, Shen L, Lehman LwH, Feng M, Ghassemi M, et~al.
\newblock MIMIC-III, a freely accessible critical care database.
\newblock Scientific data. 2016;3(1):1-9.

\bibitem{Gao2022-ep}
Gao Y, Caskey J, Miller T, Sharma B, Churpek M, Dligach D, et~al.. Tasks 1 and
  3 from progress note understanding suite of tasks: {SOAP} note tagging and
  problem list summarization. PhysioNet; 2022.

\bibitem{vadocument}
Department of~Veterans~Affairs.
\newblock VistA Mental Health Clinical Reminder Dialog Templates. 2014.

\bibitem{rogers_2021}
Rogers S. Free soap notes templates for busy healthcare professionals; 2021.
\newblock Available from:
  \url{https://blog.capterra.com/free-soap-notes-templates/}.

\bibitem{artstein2008inter}
Artstein R, Poesio M.
\newblock Inter-coder agreement for computational linguistics.
\newblock Computational linguistics. 2008;34(4):555-96.

\bibitem{eyre2021medspacy}
Eyre H, Chapman AB, Peterson KS, Shi J, Alba PR, Jones MM, et~al.
\newblock Launching into clinical space with medspaCy: a new clinical text
  processing toolkit in Python.
\newblock In: AMIA Annual Symposium Proceedings 2021; (in press, n.d.).
  Available from: \url{http://arxiv.org/abs/2106.07799}.

\bibitem{huang2015bidirectional}
Huang Z, Xu W, Yu K.
\newblock Bidirectional LSTM-CRF models for sequence tagging.
\newblock arXiv preprint arXiv:150801991. 2015.

\bibitem{yang1999re}
Yang Y, Liu X.
\newblock A re-examination of text categorization methods.
\newblock In: Proceedings of the 22nd annual international ACM SIGIR conference
  on Research and development in information retrieval; 1999. p. 42-9.

\bibitem{goldman1992revising}
Goldman HH, Skodol AE, Lave TR.
\newblock Revising axis V for DSM-IV: a review of measures of social
  functioning.
\newblock Am J Psychiatry. 1992;149:9.

\bibitem{dancey2007statistics}
Dancey CP, Reidy J.
\newblock Statistics without maths for psychology.
\newblock Pearson education; 2007.

\bibitem{sun2021nsp}
Sun Y, Zheng Y, Hao C, Qiu H.
\newblock NSP-BERT: A Prompt-based Zero-Shot Learner Through an Original
  Pre-training Task--Next Sentence Prediction.
\newblock arXiv e-prints. 2021:arXiv-2109.

\bibitem{liu2021pre}
Liu P, Yuan W, Fu J, Jiang Z, Hayashi H, Neubig G.
\newblock Pre-train, prompt, and predict: A systematic survey of prompting
  methods in natural language processing.
\newblock arXiv preprint arXiv:210713586. 2021.

\bibitem{kim2017method}
Kim J, Kwon S, Ko Y, Seo J.
\newblock A method to generate a machine-labeled data for biomedical named
  entity recognition with various sub-domains.
\newblock In: Proceedings of the International Workshop on Digital Disease
  Detection using Social Media 2017 (DDDSM-2017); 2017. p. 47-51.

\bibitem{gurulingappa2016semi}
Gurulingappa H, Toldo L, Schepers C, Bauer A, Megaro G.
\newblock Semi-Supervised Information Retrieval System for Clinical Decision
  Support.
\newblock In: TREC; 2016. .

\end{thebibliography}

\end{document}